# Circular-Line Trajectory Tracking Controller for Mobile Robot using Multi-Pixy2 Sensors

† Both authors contribute equally to this manuscript

Xuan Quang Ngo†
National Key Laboratory of Digital Control and System Engineering (DCSELab), Faculty of Mechanical Engineering, Ho Chi Minh City University of Technology (HCMUT), 268 Ly Thuong Kiet Street, District 10, Ho Chi Minh City, Vietnam.
Vietnam National University Ho Chi Minh City, Linh Trung Ward, Thu Duc City, Ho Chi Minh City, Vietnam
https://orcid.org/0000-0003-1557-5758

Tri Duc Tran†
National Key Laboratory of Digital Control and System Engineering (DCSELab), Faculty of Mechanical Engineering, Ho Chi Minh City University of Technology (HCMUT), 268 Ly Thuong Kiet Street, District 10, Ho Chi Minh City, Vietnam.
Vietnam National University Ho Chi Minh City, Linh Trung Ward, Thu Duc City, Ho Chi Minh City, Vietnam
https://orcid.org/0000-0001-6692-1943

Huy Hung Nguyen
*Faculty of Electronics and Telecommunication, Saigon University*
Ho Chi Minh, Viet Nam
https://orcid.org/0000-0002-4779-8887

Van Dong Nguyen
National Key Laboratory of Digital Control and System Engineering (DCSELab), Faculty of Mechanical Engineering, Ho Chi Minh City University of Technology (HCMUT), 268 Ly Thuong Kiet Street, District 10, Ho Chi Minh City, Vietnam.
Vietnam National University Ho Chi Minh City, Linh Trung Ward, Thu Duc City, Ho Chi Minh City, Vietnam
https://orcid.org/0000-0001-5510-9693

Van Tu Duong
National Key Laboratory of Digital Control and System Engineering (DCSELab), Faculty of Mechanical Engineering, Ho Chi Minh City University of Technology (HCMUT), 268 Ly Thuong Kiet Street, District 10, Ho Chi Minh City, Vietnam.
Vietnam National University Ho Chi Minh City, Linh Trung Ward, Thu Duc City, Ho Chi Minh City, Vietnam
https://orcid.org/0000-0002-0801-7013

Tan Tien Nguyen*
National Key Laboratory of Digital Control and System Engineering (DCSELab), Faculty of Mechanical Engineering, Ho Chi Minh City University of Technology (HCMUT), 268 Ly Thuong Kiet Street, District 10, Ho Chi Minh City, Vietnam.
Vietnam National University Ho Chi Minh City, Linh Trung Ward, Thu Duc City, Ho Chi Minh City, Vietnam
https://orcid.org/0000-0001-7943-1994

*Abstract*— This study suggests a novel tracking method that employs three Pixy2 sensors to identify the desired line trajectories instead of traditional perceiving means. Firstly, the kinematic model of the mobile robot is derived from the information gathered by three Pixy2 sensors. Secondly, the sliding mode controller is implemented to regulate the tracking error. Finally, simulation results are analyzed to show the effectiveness of the proposed method.

*Keywords*— *mobile robot, line tracking controller, sliding mode control, Pixy2 sensor-based tracking,*

## I. Introduction

The AGV (Autonomous Guided Vehicle) was the first mobile robot invented in 1953 by Barret Electronics. Since then, the mobile robot has been utilized widely in warehouses and factory as it increases efficiency, reduce labor costs, and improve flexibility in path planning [1], [2]. To guide the mobile robot, electromagnetic and optical sensors have been employed in the literature [3]. Laser sensor has also been implemented along with the Potential Field method and Enhanced Vector Field Histogram algorithms to detect obstacles [4]. Odometry, which is also known as dead reckoning, is a method for localization [5]. Localization is one of the significant tasks for autonomous robot navigation. It is one of the well-known, simple, and low-cost approaches for estimating a relative position from initial point information. Optical encoders are attached to both driving wheels and measure the wheel's angular velocity. Consequently, the position and orientation of the mobile robot are calculated. The encoders feed discretized wheel increment information to a processor, which updates the mobile robot's state continually. However, with time, odometry localization accumulates errors due to systematic errors and non-systematic errors [6]. This leads to an inaccurate estimate of the mobile robot's pose. For example, during traveling over slippery floors or fast turning, the associated encoders still register wheel revolutions though these revolutions are not converted to the linear displacement of the wheel, which leads to unbounded errors. Therefore, this solution is only suitable for short-term traveling.

Computer vision technique is a popular choice for line-following mobile robots where cameras can be utilized to detect the line trajectory [7], [8], [9]. The CMU camera is developed at Carnegie Mellon University [10] that can provide a real-time object-tracking vision system and is feasible to interface with microcontrollers and personal computers. Pixy2 sensor with a built-in CMU camera and microprocessor is a potential solution for the line-follower mobile robot [11]. This approach is low cost and power consumption. The open Pixy2' library provides the line-tracking algorithm, which is able to output the coordinate of a center point of the detected line segment quickly.

In our previous study, to effectively deal with the junction line, an error model for the mobile robot using one Pixy2 sensor has been derived [11]. However, the utilization of a single sensor installed on the mobile robot as a means of perceiving suffers from the following drawbacks. Firstly, although the mobile robot can track the line well, its position in the global world coordinate is unknown. Secondly, when the mobile robot turns at high velocity, its inertia makes the mobile robot out of the line trajectory. Thirdly, in the case when the sensor is broken, or the line is impaired; the line trajectory tracking process is discontinued, and thus the mobile robot should be stopped. To alleviate the aforementioned problems, the tracking controller is proposed based on the novel model using three Pixy2 sensors for the mobile robot following the circle or the curve. These solutions can bring us some following advantages.

- Based on the center of the line curve, which is estimated by three Pixy2 sensors, the precise position of the mobile robot in the global world can be achieved.

- The centers of the detected line segment are available at any time interval due to using three Pixy2 sensors. As a result, the center and radius of the curve can be calculated. It facilitates the detection of all points of the curve at any time interval. Therefore, when the curve is not visible in the sensor frame, the mobile robot still can return to the curve by using other methods such as odometry.

- Once a sensor is not reliable due to a scratch or faded line, the mobile robot still can complete its task based on the data of the sensors left.

This paper is organized as follows. Section II proposes a tracking error model of the mobile robot, which is derived from the data from three Pixy2 sensors. Then, in section III, a tracking controller is constructed based on the tracking error to control the mobile robot. Finally, section IV shows simulation results and a detailed analysis of the proposed controller.

## II. MODELING A MOBILE ROBOT USING MULTI-PIXY2 SENSORS

The Pixy2 sensor with the sensor frame size $w \times l$ (mm), is used to detect the center of the captured line segment point $C({}^Cx_C, {}^Cy_C)$, point $A({}^Cx_A, {}^Cy_A)$, and point $B({}^Cx_B, {}^Cy_B)$ superscript stands for Pixy2 sensor coordinate $\{C\}$ as shown in **Error! Reference source not found.**. Due to the limited sensor frame, $0 \leq {}^Cx_i \leq w$ and $0 \leq {}^Cy_i \leq l$., where $i = A, B, C$.

Similarly, it can infer that sensor 1, sensor 2, and sensor 3 as shown in Fig. 2, the center of three captured line segments $({}^{C1}x_{C1}, {}^{C1}y_{C1}), ({}^{C2}x_{C2}, {}^{C2}y_{C2}), ({}^{C3}x_{C3}, {}^{C3}y_{C3})$ in the sensor coordinate $\{C_1\}, \{C_2\}, \{C_3\}$ respectively is detected. Three sensors are mounted on the mobile robot with the distance of the origin of the sensor coordinate $\{C_1\}, \{C_2\}, \{C_3\}$ with respect to (w.r.t) the mobile robot coordinate $\{R\}$ as shown in Fig. 2.

Given the description of the position vector ${}^RP$ relative to the mobile robot coordinate $\{R\}$, the description of the position vector ${}^GP$ relative to the global coordinate $\{G\}$ can be expressed as follows [12]:

$$^GP = {}^G_RR \ {}^RP + {}^GP_{RORG} \qquad (1)$$

where, ${}^G_RR$ is the rotation matrix describing $\{R\}$ w.r.t $\{G\}$; ${}^GP_{RORG}$ is the vector that locates the origin of $\{R\}$ w.r.t $\{G\}$.

From (1), the center of the captured line segment detected by sensor 1, 2, and 3 w.r.t $\{R\}$ are given as follows:

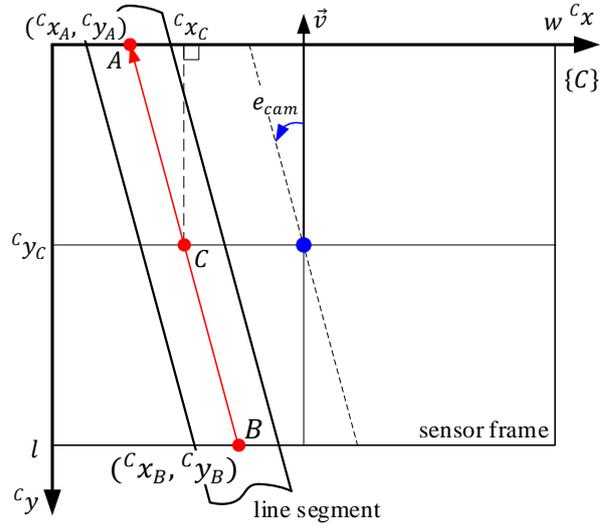

Fig. 1 The Pixy2 sensor coordinate $\{C\}$

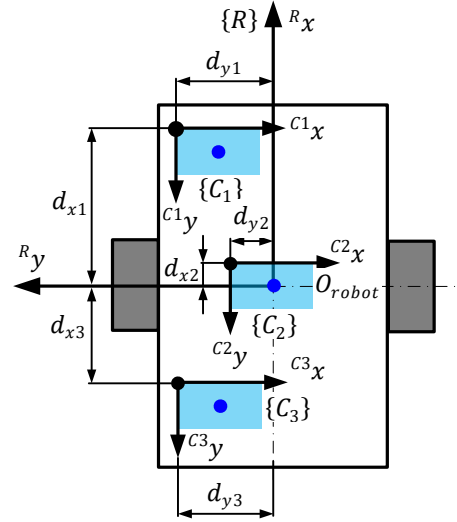

Fig. 2 The mobile robot structure using three Pixy2 sensors

$$\begin{bmatrix} {}^Rx_{C1} \\ {}^Ry_{C1} \end{bmatrix} = \begin{bmatrix} \cos\frac{\pi}{2} & \sin\frac{\pi}{2} \\ -\sin\frac{\pi}{2} & \cos\frac{\pi}{2} \end{bmatrix} \begin{bmatrix} 1 & 0 \\ 0 & -1 \end{bmatrix} \begin{bmatrix} {}^{C1}x_{C1} \\ {}^{C1}y_{C1} \end{bmatrix} + \begin{bmatrix} d_{x1} \\ d_{y1} \end{bmatrix} \qquad (2)$$

$$\begin{bmatrix} {}^Rx_{C2} \\ {}^Ry_{C2} \end{bmatrix} = \begin{bmatrix} \cos\frac{\pi}{2} & \sin\frac{\pi}{2} \\ -\sin\frac{\pi}{2} & \cos\frac{\pi}{2} \end{bmatrix} \begin{bmatrix} 1 & 0 \\ 0 & -1 \end{bmatrix} \begin{bmatrix} {}^{C2}x_{C2} \\ {}^{C2}y_{C2} \end{bmatrix} + \begin{bmatrix} d_{x2} \\ d_{y2} \end{bmatrix} \qquad (3)$$

$$\begin{bmatrix} {}^Rx_{C3} \\ {}^Ry_{C3} \end{bmatrix} = \begin{bmatrix} \cos\frac{\pi}{2} & \sin\frac{\pi}{2} \\ -\sin\frac{\pi}{2} & \cos\frac{\pi}{2} \end{bmatrix} \begin{bmatrix} 1 & 0 \\ 0 & -1 \end{bmatrix} \begin{bmatrix} {}^{C3}x_{C3} \\ {}^{C3}y_{C3} \end{bmatrix} + \begin{bmatrix} -d_{x3} \\ d_{y3} \end{bmatrix} \qquad (4)$$

where, $d_{xi}$ and $d_{yi}$ are the distance from the origin of $\{C_i\}$ to $O_{robot}{}^Ry$ axis and $O_{robot}{}^Rx$ axis respectively.

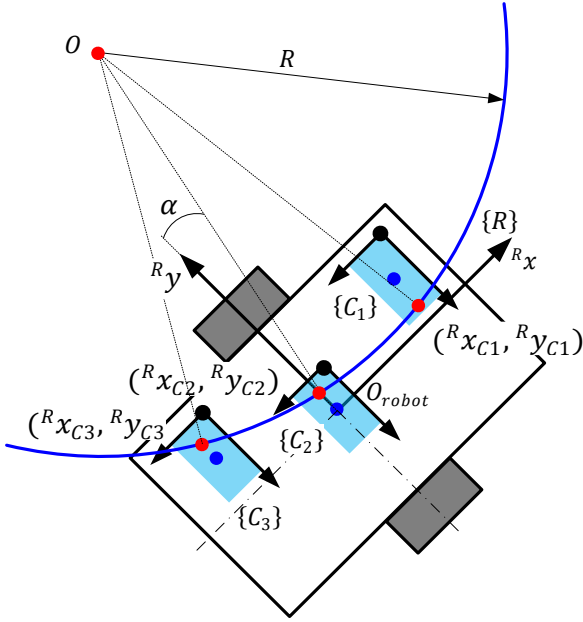

Fig. 3 The center of the curve w.r.t the mobile robot coordinate $\{R\}$

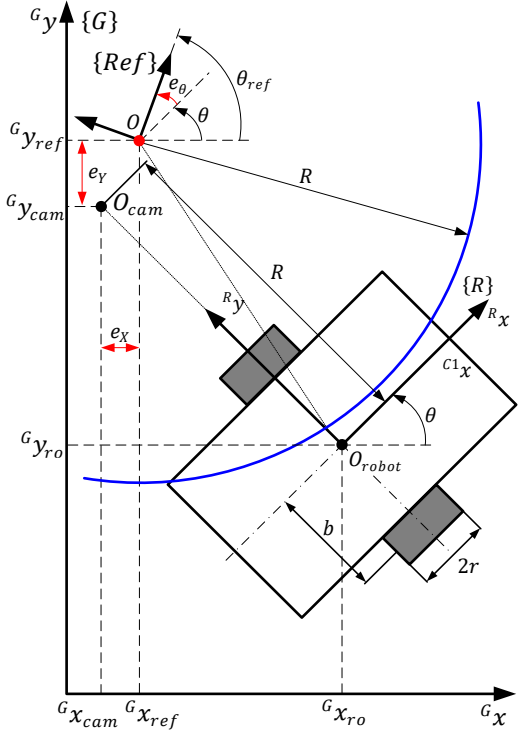

Fig. 4 The tracking errors of the mobile robot

Let us define the point $O$ as the center of the curve with radius $R$ passing through three points $(^Rx_{C1}, {}^Ry_{C1}), (^Rx_{C2}, {}^Ry_{C2}), (^Rx_{C3}, {}^Ry_{C3})$, it is coordinate $(^Rx_{line}, {}^Ry_{line})$ w.r.t $\{R\}$ as shown in Fig. 3 can be obtained as:

$$\begin{bmatrix}^Rx_{line}\\{}^Ry_{line}\end{bmatrix} = \begin{bmatrix}2(^Rx_{C2} - {}^Rx_{C1}) & 2(^Ry_{C2} - {}^Ry_{C1})\\2(^Rx_{C3} - {}^Rx_{C1}) & 2(^Ry_{C3} - {}^Ry_{C1})\end{bmatrix}$$
$$\begin{bmatrix}-{}^Rx_{C1}^2 + {}^Rx_{C2}^2 - {}^Ry_{C1}^2 + {}^Ry_{C2}^2\\-{}^Rx_{C1}^2 + {}^Rx_{C3}^2 - {}^Ry_{C1}^2 + {}^Ry_{C3}^2\end{bmatrix} \quad (5)$$

**Assumption 1.** Assume that at the initial time, it is ensured that three sensors capture the line segment as shown in Fig. 1, and the mobile robot's initial center point is $(^Gx_{ro0}, {}^Gy_{ro0})$ w.r.t $\{G\}$ and the initial heading angle of the mobile robot is $\theta_0$ w.r.t $\{G\}$.

**Assumption 2.** Assume that at the initial time, the center of the captured line segment detected by sensor 1, 2, and 3 w.r.t $\{R\}$ are $(^Rx_{C10}, {}^Ry_{C10}), (^Rx_{C20}, {}^Ry_{C20}), (^Rx_{C30}, {}^Ry_{C30})$. Substituting these points into (5) yields the center of the curve $(^Rx_{line0}, {}^Ry_{line0})$ w.r.t $\{R\}$ at the initial time.

Based on (1) and the relationship between $\{R\}$ and $\{G\}$ as shown in Fig. 4, the center of the curve w.r.t $\{G\}$ can be obtained as follows:

$$\begin{bmatrix}^Gx_{ref}\\{}^Gy_{ref}\end{bmatrix} = \begin{bmatrix}\cos\theta_0 & -\sin\theta_0\\\sin\theta_0 & \cos\theta_0\end{bmatrix}\begin{bmatrix}^Rx_{line0}\\{}^Ry_{line0}\end{bmatrix} + \begin{bmatrix}^Gx_{ro0}\\{}^Gy_{ro0}\end{bmatrix} \quad (6)$$

**Assumption 3.** Assume that sensor 1, 2, and 3 are installed on the mobile robot in such a way that the centers of the three sensors lenses create the curve with radius $R$, and the center of sensor 2's lens coincides with the center of the mobile robot. As a result, let us define $O_{cam}$ having $(0, R)$ w.r.t $\{R\}$ and $(^Gx_{cam}, {}^Gy_{cam})$ w.r.t $\{G\}$ as shown in Fig. 4.

The coordinate of the tracking point $O_{cam}$ w.r.t $\{G\}$:

$$\begin{bmatrix}^Gx_{cam}\\{}^Gy_{cam}\end{bmatrix} = \begin{bmatrix}^Gx_{ref} - R\sin\theta\\{}^Gy_{ref} + R\cos\theta\end{bmatrix} - \begin{bmatrix}\cos\theta & -\sin\theta\\\sin\theta & \cos\theta\end{bmatrix}\begin{bmatrix}^Rx_{line}\\{}^Ry_{line}\end{bmatrix} \quad (7)$$

where, $\theta$ is the heading angle of the mobile robot, defined by (8); $R$ is the radius of the curve.

Let us define the reference coordinate $\{Ref\}$, its origin is located at the point $(^Gx_{ref}, {}^Gy_{ref})$ and the angle between $\{Ref\}$ and $\{G\}$ is $\theta_{ref}$ as shown in Fig. 4.

The heading angle of the mobile robot w.r.t $\{G\}$ is obtained as follows:

$$\theta = \theta_{ref} - e_{cam} \quad (8)$$

where, $e_{cam}$ is the angle error w.r.t $\{C_2\}$ between the line segment captured by sensor 2 and the mobile robot heading as shown in Fig. 1, defined by (9).

The angle error between the line segment captured by sensor 2 and the mobile robot heading w.r.t $\{C_2\}$:

$$e_{cam} = \operatorname{atan}\left(\frac{^{C2}x_{B2} - {}^{C2}x_{A2}}{^{C2}y_{B2} - {}^{C2}y_{A2}}\right) \quad (9)$$

where, $(^{C2}x_{A2}, {}^{C2}y_{A2}), (^{C2}x_{B2}, {}^{C2}y_{B2})$ are the coordinate of point $A$ and point $B$ w.r.t $\{C_2\}$, collected by sensor 2.

Based on the ordinary kinematic model of a mobile robot with two actuated wheels [11]. The dynamic equation of mobile robot at the tracking point $O_{cam}$:

$$\begin{bmatrix}^G\dot{x}_{cam}\\{}^G\dot{y}_{cam}\\\dot{\theta}\end{bmatrix} = \begin{bmatrix}\cos\theta & -R\cos\theta\\\sin\theta & -R\sin\theta\\0 & 1\end{bmatrix}\begin{bmatrix}v\\w\end{bmatrix} \quad (10)$$

where, $v$ and $w$ are the linear and angular velocities of the mobile robot.

The relationship between $v, w$ and the angular velocities of the two-driving wheel is [13]:

$$\begin{bmatrix} w_{rw} \\ w_{lw} \end{bmatrix} = \begin{bmatrix} 1/r & b/r \\ 1/r & -b/r \end{bmatrix} \begin{bmatrix} v \\ w \end{bmatrix} \quad (11)$$

where, $w_{rw}, w_{lw}$ are the angular velocities of the right and left wheels; $b$ is the distance from the mobile robot's center point to the driving wheel; $r$ is the driving wheel radius.

By defining the tracking errors, as the difference between the mobile robot coordinate $\{R\}$ when it is transformed to point $O_{cam}$ and the reference coordinate $\{Ref\}$ as $e_x \triangleq {}^G x_{ref} - {}^G x_{cam}; e_y \triangleq {}^G y_{ref} - {}^G y_{cam}; e_\theta \triangleq \theta_{ref} - \theta$ in $\{G\}$, the tracking errors w.r.t $\{R\}$ are given as follows:

$$\begin{bmatrix} e_1 \\ e_2 \\ e_3 \end{bmatrix} = \begin{bmatrix} \cos\theta & \sin\theta & 0 \\ -\sin\theta & \cos\theta & 0 \\ 0 & 0 & 1 \end{bmatrix} \begin{bmatrix} e_x \\ e_y \\ e_\theta \end{bmatrix} \quad (12)$$

The dynamic of the tracking errors is given as follows:

$$\begin{bmatrix} \dot{e}_1 \\ \dot{e}_2 \\ \dot{e}_3 \end{bmatrix} = \begin{bmatrix} 0 \\ 0 \\ w_{ref} \end{bmatrix} + \begin{bmatrix} -1 & e_2 + R \\ 0 & -e_1 \\ 0 & -1 \end{bmatrix} \begin{bmatrix} v \\ w \end{bmatrix} \quad (13)$$

## III. CONTROLLER DESIGN

In order to guarantee the convergence of all the tracking errors $\boldsymbol{e} = [e_1 \quad e_2 \quad e_3]^T$, let us define the sliding surface for the system of (13) as follows:

$$\boldsymbol{s} = \begin{bmatrix} s_1 \\ s_2 \\ s_3 \end{bmatrix} = \begin{bmatrix} e_1 + e_2 \\ e_1 - e_2 \\ -e_1 + e_2 + e_3 \end{bmatrix} \quad (14)$$

Taking the first-time derivative of the sliding surface, it yields:

$$\dot{\boldsymbol{s}} = \boldsymbol{f}(\boldsymbol{e}) + \boldsymbol{g}(\boldsymbol{e})\boldsymbol{u}_v \quad (15)$$

where, $\boldsymbol{g}(\boldsymbol{e}) = \begin{bmatrix} -1 & e_2 + R - e_1 \\ -1 & e_2 + R + e_1 \\ 1 & -(e_2 + R + e_1 + 1) \end{bmatrix}$;

$\boldsymbol{f}(\boldsymbol{e}) = \begin{bmatrix} 0 \\ 0 \\ w_{ref} \end{bmatrix}; \boldsymbol{u}_v = \begin{bmatrix} v \\ w \end{bmatrix}; \boldsymbol{e} = \begin{bmatrix} e_1 \\ e_2 \\ e_3 \end{bmatrix};$

Now, the Lyapunov function is chosen as:

$$V(\boldsymbol{s}) = \frac{1}{2} \boldsymbol{s}^T \boldsymbol{s} \geq 0 \quad (16)$$

and its derivative is:

$$\dot{V}(\boldsymbol{s}) = \boldsymbol{s}^T \dot{\boldsymbol{s}} = \boldsymbol{s}^T (\boldsymbol{f}(\boldsymbol{e}) + \boldsymbol{g}(\boldsymbol{e})\boldsymbol{u}_v) \quad (17)$$

The tracking control law is as follows:

$$\boldsymbol{u}_v = -\boldsymbol{g}(\boldsymbol{e})^{-1}(\boldsymbol{f}(\boldsymbol{e}) + K\, sign(\boldsymbol{s})) \quad (18)$$

where, $K$ is the positive constant.

Substituting (18) into (17), one achieves $\dot{V}(\boldsymbol{s}) = -k|s|$. Then, it can be seen that the derivative of the selected Lyapunov function is negative definite. Following the Lyapunov stability theorem [14], the sliding surface $\boldsymbol{s} = 0$ is asymptotically stable. As a result, the tracking error $\boldsymbol{e} \to 0$ as $t \to \infty$.

Chattering characterizes negative effects on the controllers, so it should be eliminated. This can be achieved by substituting the sign function of (18) with the saturation function. This allows for smoothing out the control discontinuity in a thin boundary layer neighboring the switching surface [15].

The saturation function is defined as [15]:

$$\begin{cases} sat(s) = s & , if\ |s| \leq 1 \\ sat(s) = sign(s) & , otherwise \end{cases} \quad (19)$$

## IV. SIMULATION RESULTS AND DISCUSSION

To verify the effectiveness of the proposed model and tracking controller, the following simulations have been done for the wheel mobile robot with a defined circle. The numerical values used in this simulation are given in TABLE I.

TABLE I. THE NUMERICAL VALUES FOR SIMULATIONS

| Parameters | Value | Unit |
|---|---|---|
| $b$ | 0.365 | $m$ |
| $r$ | 0.047 | $m$ |
| $R$ | 1 | $m$ |
| $w_{ref}$ | 1 | $rad/s$ |
| ${}^G x_{ref}$ | 2 | $m$ |
| ${}^G y_{ref}$ | 2 | $m$ |

The initial position and heading angle of the mobile robot are ${}^G x_{ro} = 1.8; {}^G y_{ro} = 0.8$ and $\theta = 1^o$. The gains of the controller are $k_1 = 0.01, k_2 = 0.5, k_3 = 1$. As shown in Fig. 5, the trajectories of the tracking point $O_{cam}$ follows a curved trajectory toward the center of the circle at the coordinate $O_{ref}(2,2)$, and the center of the mobile robot follows the desired circle consequently. It can be observed that the tracking errors go to zeros after about $6\ s$ as shown in Fig. 6. The simulations of wheel velocities with smoothing are shown in Fig. 7.

To describe how the change in the controller's gains affects the trajectory and response of the mobile robot, the second simulation is conducted. The mobile robot is placed outside the circle, namely ${}^G x_{ro} = 1.8; {}^G y_{ro} = 0.8$ as shown in Fig. 8. With the gains of the controllers chosen as $k_1 = 0.05, k_2 = 0.2, k_3 = 1$, the ratio of the linear velocity and the

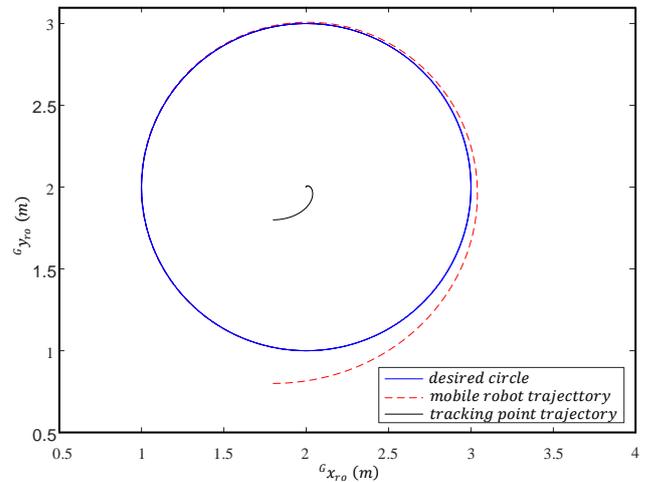

Fig. 5 First scenario - Trajectories of the tracking point and the center of the mobile robot

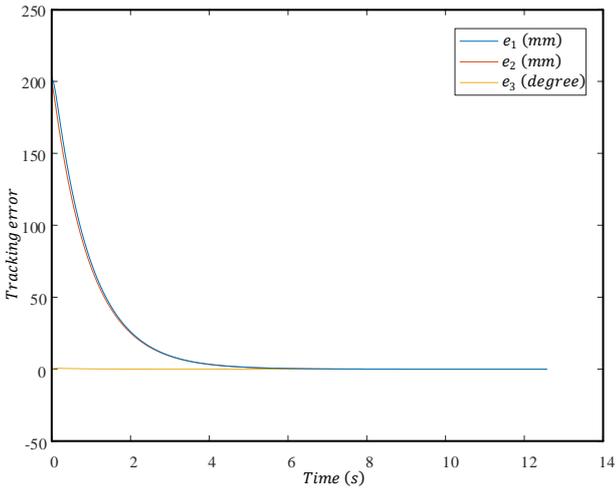

Fig. 6 First scenario - The tracking error

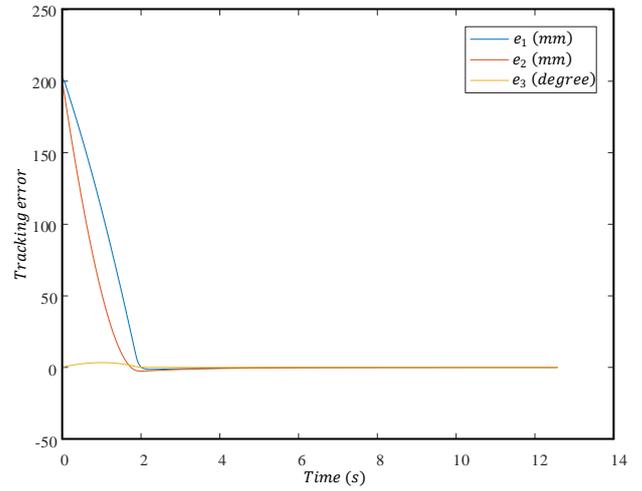

Fig. 9 Second scenario - The tracking error for $k_1 = 0.05; k_2 = 0.2; k_3 = 1$

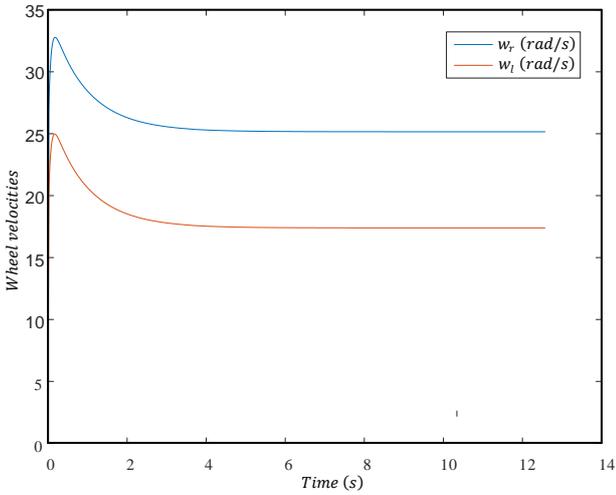

Fig. 7 First scenario - The wheel velocities with smoothing

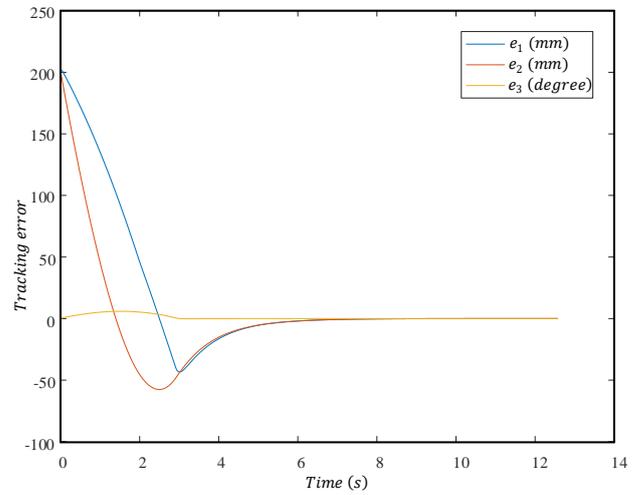

Fig. 10 Second scenario - The tracking error for $k_1 = 0.01; k_2 = 0.2; k_3 = 1$

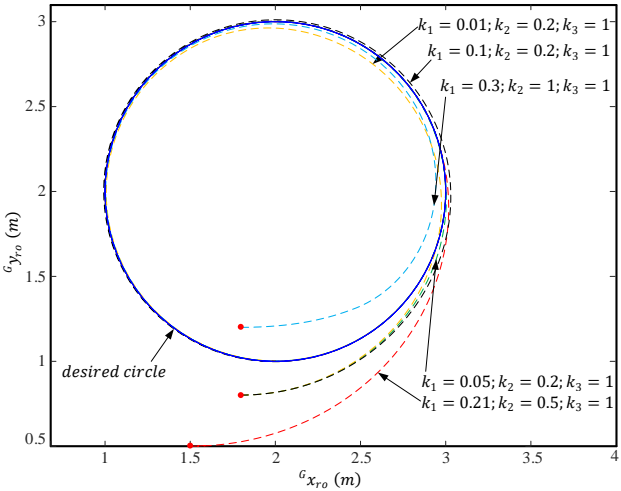

Fig. 8 Second scenario - The mobile robot trajectory with the change of the controller's gains and different initial positions

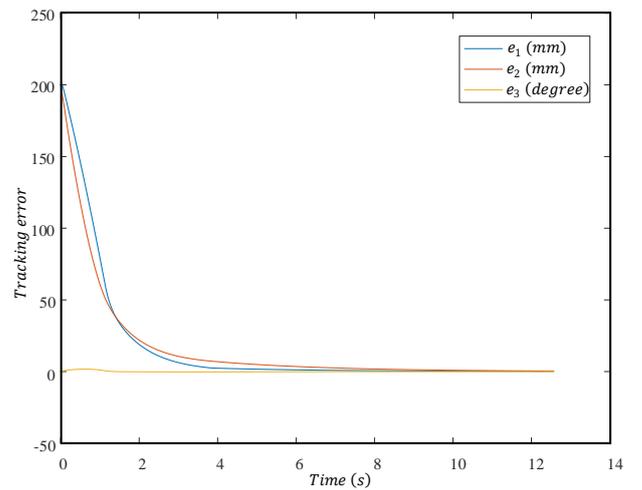

Fig. 11 Second scenario - The tracking error for $k_1 = 0.1; k_2 = 0.2; k_3 = 1$

angular velocity is large enough to drive the mobile robot to the circle quickly as shown in Fig. 12, and the tracking errors converge to zero after about $3\ s$ as shown in Fig. 9. If the parameters of the controllers are $k_1 = 0.01; k_2 = 0.2, k_3 = 1$ leading to the ratio is smaller than the adequate ratio, the mobile robot trajectory will be as shown in Fig. 8, and it will take about $8\ s$ for tracking errors to converge to zero as shown in Fig. 10. Similarly, if the parameters of the controllers are $k_1 = 0.1; k_2 = 0.2, k_3 = 1$ leading to the ratio is larger than the adequate ratio, the mobile robot trajectory will be as shown in Fig. 8, and it will take about $6\ s$ for tracking errors to go to zero as shown in Fig. 11.

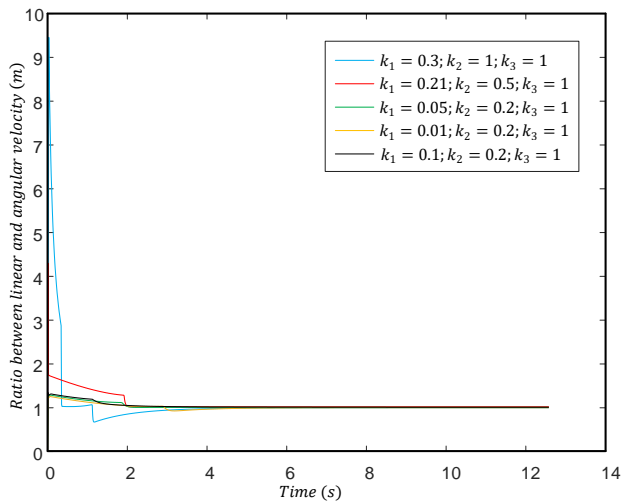

Fig. 12 The second simulation - The ratio between the linear velocity and angular velocity

When the mobile robot starts from inside the circle, namely $^Gx_{ro} = 1.8$; $^Gy_{ro} = 1.2$, the mobile robot follows the circle with the parameters chosen as $k_1 = 0.3, k_2 = 1, k_3 = 1$ as shown in Fig. 8. To make the mobile robot track the circle quickly, the ratio should be large as shown in Fig. 12. As a result, the mobile robot's trajectory is a straight line at the beginning, and then slowly curve toward the desired circle and finally tangentially move along the desired circle as shown in Fig. 8.

When the mobile robot is placed far outside the circle, namely $^Gx_{ro} = 1.5$; $^Gy_{ro} = 0.5$, the mobile robot is still able to track the circle with the parameters chosen as $k_1 = 0.21, k_2 = 0.5, k_3 = 1$. The ratio should be large enough for the mobile robot to embrace the desired circle gradually as shown in Fig. 8 and Fig. 12.

Because of the model characteristics, the gains of the controllers should be chosen in such a way that the ratio of the linear velocity and the angular velocity is large enough for the mobile robot to track the circle as quickly as possible. If the gains are quite large, the chattering phenomenon will appear and ultimately degrade the control performance. If the gains are quite small, it spends much time tracking the circle.

ACKNOWLEDGMENT

The research is funded by Vietnam National University Ho Chi Minh City (VNU-HCM) under grant number TX2023-20b-01. We acknowledge the support of time and facilities from National Key Laboratory of Digital Control and System Engineering (DCSELab), Ho Chi Minh University of Technology (HCMUT), VNU-HCM for this study.